\documentclass[11pt]{article} 
\usepackage[table]{xcolor}
\usepackage{rldmsubmit,palatino}
\usepackage{graphicx}
\usepackage{amsmath}
\usepackage{amssymb}
\usepackage{bm}
\RequirePackage{natbib}     
\usepackage{fontawesome}

\usepackage{fvextra}
\usepackage{listings}
\usepackage{caption}
\usepackage{fancyvrb}
\usepackage{todonotes}
\usepackage{algorithm}
\usepackage{algpseudocode}
\usepackage{subcaption}
\usepackage{hyperref}       

\newcommand{\y}{\gamma}
\newcommand{\s}{\sigma}

\newcommand{\N}{\mathcal{N}}

\newcommand{\B}{\mathcal{B}}

\DeclareMathOperator*{\argmin}{argmin}

\DeclareMathOperator*{\clip}{clip}

\definecolor{star_yellow}{HTML}{FFD700}
\definecolor{cas_blue}{HTML}{0014C7}
\definecolor{casit_red}{HTML}{C12A22}
\definecolor{liauto_blue}{HTML}{225450}
\newcommand{\pc}{{\color{casit_red}\boldsymbol{c}}}
\newcommand{\pu}{{\color{cas_blue}\boldsymbol{u}}}
\newcommand{\pl}{{\color{liauto_blue}\boldsymbol{l}}}

\title{CAMEL: Continuous Action Masking Enabled by Large Language Models for Reinforcement Learning}

\author{
Yanxiao Zhao$^{\pc\hspace{.1em}\pu\hspace{.1em}\pl}$ \quad
Yangge Qian$^{\hspace{.1em}\pc\hspace{.1em}\pu}$ \quad
Jingyang Shan$^{\hspace{.1em}\pc\hspace{.1em}\pu}$ \quad
Xiaolin Qin$^{\pc\hspace{.1em}\pu}$ \\
\texttt{\{zhaoyanxiao21, qianyange20, shanjingyang21\}@mails.ucas.ac.cn}\\
\texttt{qinxl2001@126.com}\\
$^{\pc}$ Chengdu Institute of Computer Applications, Chinese Academy of Sciences \\
$^{\pu}$ School of Computer Science and Technology, University of Chinese Academy of Sciences \\
$^{\pl}$ Li Auto \\
}

\begin{document}

\maketitle

\begin{abstract}
Reinforcement learning (RL) in continuous action spaces encounters persistent challenges, such as inefficient exploration and convergence to suboptimal solutions. To address these limitations, we propose CAMEL (Continuous Action Masking Enabled by Large Language Models), a novel framework integrating LLM-generated suboptimal policies into the RL training pipeline. CAMEL leverages dynamic action masking and an adaptive epsilon-masking mechanism to guide exploration during early training stages while gradually enabling agents to optimize policies independently.
At the core of CAMEL lies the integration of Python-executable suboptimal policies generated by LLMs based on environment descriptions and task objectives. Although simplistic and hard-coded, these policies offer valuable initial guidance for RL agents. To effectively utilize these priors, CAMEL employs masking-aware optimization to dynamically constrain the action space based on LLM outputs. Additionally, epsilon-masking gradually reduces reliance on LLM-generated guidance, enabling agents to transition from constrained exploration to autonomous policy refinement.
Experimental validation on Gymnasium MuJoCo environments (Hopper-v4, Walker2d-v4, Ant-v4) demonstrates the effectiveness of CAMEL. In Hopper-v4 and Ant-v4, LLM-generated policies significantly improve sample efficiency, achieving performance comparable to or surpassing expert masking baselines. For Walker2d-v4, where LLMs struggle to accurately model bipedal gait dynamics, CAMEL maintains robust RL performance without notable degradation, highlighting the framework’s adaptability across diverse tasks.
While CAMEL shows promise in enhancing sample efficiency and mitigating convergence challenges, these issues remain open for further research. Future work aims to generalize CAMEL to multimodal LLMs for broader observation-action spaces and automate policy evaluation, reducing human intervention and enhancing scalability in RL training pipelines.
\end{abstract}

\keywords{Large Language Models, Reinforcement Learning,\\LLMs Enhanced RL, Action Masking}

\acknowledgements{
This research was partly supported by the Sichuan Science and Technology Program (2024NSFJQ0035, 2024NSFSC0004 ), and the Talents by Sichuan provincial Party Committee Organization Department.}

\startmain 

\section{Introduction}

Large Language Models (LLMs), such as OpenAI o1 and Google Gemini, have demonstrated remarkable capabilities in reasoning and code generation.
These advancements have spurred increasing interest in applying LLMs to decision-making tasks. However, decision-making often requires not only reasoning and prior knowledge but also the ability to adapt and learn interactively—a hallmark of Reinforcement Learning (RL). RL achieves this by enabling agents to interact with their environment, observe feedback in the form of rewards, and iteratively refine their policies.
This interactive learning mechanism has driven RL's success in various domains, including robotics, strategic gaming, and autonomous control.
The integration of RL with LLMs promises to combine their complementary strengths, paving the way for more sample-efficient learning and enhanced decision-making performance.

While significant progress has been made in leveraging LLMs to augment various RL components—serving as reward designers, information processors, or world model simulators~\citep{cao2024survey}—the potential of LLMs as expert policies to directly guide RL agents remains largely unexplored.
This underutilization arises from two key challenges: first, the suboptimal performance of LLM-based policies without specialized fine-tuning; and second, the inherent vulnerability of RL algorithms to convergence to suboptimal solutions when guided by imprecise or unreliable feedback. 
Previous works, such as \citet{hausknecht2020interactive} and \citet{yao2020keep}, have primarily focused on text-based game environments, where LLMs generate or refine actions. However, these approaches often suffer from domain specificity and limited generalizability to more complex RL scenarios.



\begin{figure}[ht]
    \centering
    \includegraphics[width=1.0\linewidth]{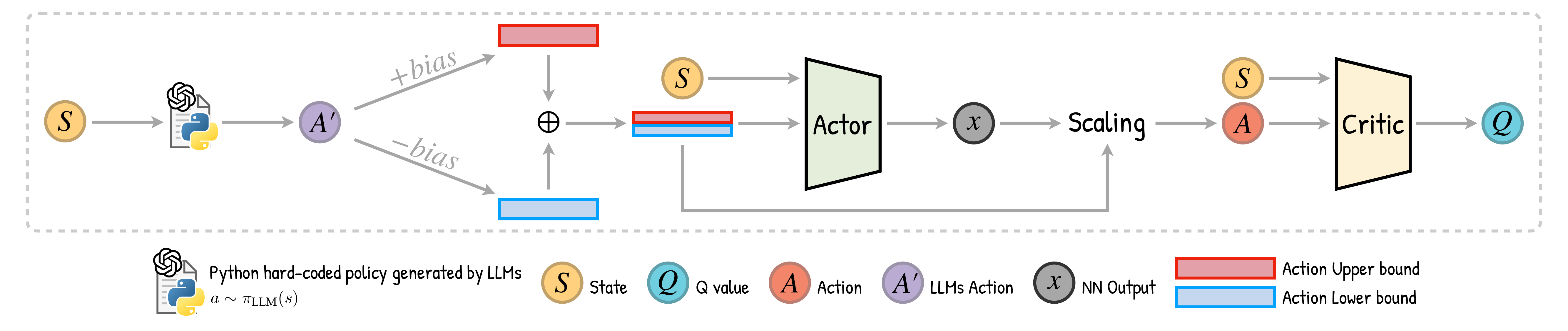}
    \caption{Diagram of the \textsc{camel} RL Model Forward Pipeline: A concise overview of the framework.}\label{fig:camel_pipeline}
\end{figure}

\textbf{Contribution.}
1. We demonstrate the potential of LLMs in controlling multi-joint robots with continuous action spaces. By generating less than $100$ lines of code, these models provide effective initial suboptimal policies for \texttt{Hopper-v4} and \texttt{Ant-v4} environments, which significantly support subsequent RL training.
2. We propose the \textsc{camel} framework, which leverages LLM-generated suboptimal policies to guide RL training by dynamically constraining the action space and employing progressive strategy optimization, significantly improving sample efficiency and effectively avoiding poor solutions. 
3. We conduct experiments on Gymnasium MuJoCo \texttt{Hopper-v4}, \texttt{Walker2d-v4}, and \texttt{Ant-v4}~\citep{todorov2012mujoco, towers_gymnasium_2023}.
In \texttt{Hopper-v4} and \texttt{Ant-v4}, LLM-generated policies significantly improve RL sample efficiency.
In \texttt{Walker2d-v4}, the LLM fails to generate effective policies, which is likely caused by challenges in modeling the complexity of bipedal gait alternation.
Nevertheless, RL agents exhibit robust performance, showing no significant improvement or degradation even under poor policy guidance.






\section{Preliminaries}

RL involves an agent interacting with an environment modeled as a Markov Decision Process (MDP) $(\mathcal{S}, \mathcal{A}, P, R, \gamma)$. 
The agent observes a state $s \in \mathcal{S}$, takes an action $a \in \mathcal{A}$ via a policy $\pi(a|s)$, transitions to $s' \sim P(s'|s, a)$, and receives a reward $r = R(s, a)$.
The objective is to learn a policy that maximizes the expected cumulative reward, $\mathbb{E}[\sum_{t=0}^\infty \gamma^t R(s_t, a_t)]$.
In this work, prior knowledge is encoded into the policy using $\pi_{\text{LLM}}$, a Python-executable function generated by an LLM.

\section{Approach}

In this section, we present a detailed introduction to the \textsc{camel} framework, with its schematic workflow illustrated in Figure~\ref{fig:camel_pipeline} and the pseudocode (Algorithm~\ref{alg:camel_td3}).
The \textsc{camel} framework consists of three key components.
1. \textit{Utilizing LLMs to Generate Hard-Coded Policies}.
This component leverages LLMs to generate Python code that encodes prior knowledge, assuming the optimal policy is near $\pi_{\text{LLM}}$ and guiding policy learning.
2. \textit{Masking-Aware Continuous Action Masking}.
In this component, masking information is incorporated into $s$ and input to the actor model, enabling it to learn and adapt to dynamic masking.
By constraining the action space based on $\pi_{\text{LLM}}$, the RL agent explores more efficiently.
3. \textit{Epsilon-Masking}.
Action masking is applied with a probability of $1-\epsilon$.
As training progresses, masking intensity decreases, allowing the RL agent to overcome dependency on $\pi_{\text{LLM}}$ and achieve higher rewards in the original environment.

\begin{algorithm}[tb]
   \caption{\textsc{camel}-TD3}
   \label{alg:camel_td3}
\begin{algorithmic}[1]
   \State Initialize critic networks $Q_{\theta_1}$, $Q_{\theta_2}$, and actor network $\pi_\phi$ with random parameters $\theta_1$, $\theta_2$, $\phi$
   \State Initialize target networks $\theta'_1 \leftarrow \theta_1$, $\theta'_2 \leftarrow \theta_2$, $\phi' \leftarrow \phi$
   \State Initialize \textcolor{red}{LLM-generated policy $a \sim \pi_{\text{LLM}}(s)$} and replay buffer $\B$   
   \For{$t=1$ {\bfseries to} $T$}
   \State Generate action bounds: \textcolor{red}{$a_{lb} = \pi_{\text{LLM}}(s) - bias$, $a_{ub} = \pi_{\text{LLM}}(s) + bias$}
   \State Apply \textcolor{red}{$\epsilon$-masking}: with probability $1-\epsilon$, set $a_{lb}, a_{ub}$ to action space bounds
   \State Select output \textcolor{red}{$x \sim \pi_\phi(s, a_{lb}, a_{ub})$}, $x \in [0, 1]$,
   \State Map output to action: $\textcolor{red}{a = \Call{actionMapping}{x, a_{lb}, a_{ub}}} + \eta$,
   \State $\eta \sim \N(0, \sigma)$ and observe reward $r$ and new state $s'$
   \State Store transition tuple $(s, a_{lb}, a_{ub}, a, r, s')$ in $\B$
   \State 
   \State Sample mini-batch of $N$ transitions $(s, a_{lb}, a_{ub}, a, r, s')$ from $\B$
   \State $\tilde a \leftarrow \Call{actionMapping}{\pi_{\phi'}(s', a_{lb}', a_{ub}'), a_{lb}', a_{ub}'} + \eta, \quad \eta \sim \clip(\N(0, \tilde \s), -c, c)$
   \State $y \leftarrow r + \y \min_{i=1,2} Q_{\theta'_i}(s', \tilde a)$
   \State Update critics $\theta_i \leftarrow \argmin_{\theta_i} N^{-1} \sum (y - Q_{\theta_i}(s,a))^2$
   \If{$t$ mod $d$}
   \State Update $\phi$ by the deterministic policy gradient:
   \State \textcolor{red}{$\nabla_{\phi} J(\phi) = N^{-1} \sum \nabla_{a} Q_{\theta_1}(s, a) |_{a=\Call{actionMapping}{\pi_{\phi}(s, a_{lb}, a_{ub}), a_{lb}, a_{ub}}} \nabla_{\phi} \Call{actionMapping}{\pi_{\phi}(s, a_{lb}, a_{ub}), a_{lb}, a_{ub}}$}
   \State Update target networks: $\theta'_i \leftarrow \tau \theta_i + (1 - \tau) \theta'_i$, $\phi' \leftarrow \tau \phi + (1 - \tau) \phi'$
   \EndIf
   \EndFor
   \Function{actionMapping}{$x, a_{lb}, a_{ub}$}
       \State $scale = \frac{a_{ub} - a_{lb}}{2}$, $bias = \frac{a_{ub} + a_{lb}}{2}$
       \State \Return $x \cdot scale + bias$
   \EndFunction
\end{algorithmic}
\end{algorithm}

\subsection{Utilizing LLMs to Generate Hard-Coded Policies}

Our prompt (see Figure~\ref{fig:llm_prompt}) includes detailed information about the state space $\mathcal{S}$ and the action space $\mathcal{A}$, clarifying the dimensions of each, the task objectives, and the MuJoCo XML file.
These resources are drawn from the Gymnasium documentation and codebase~\citep{towers_gymnasium_2023}.
The goal is to create a Python policy function that uses hard-coded parameters to map the environment state $s$ to an action $a_{\text{LLM}}$.

We adopt the Chain of Thought approach to guide the LLM with step-by-step reasoning instructions for policy design.
However, the generated parameters are hard-coded, making the policies non-adaptive to environmental feedback and potentially unstable.
To select the optimal policy, we generate multiple candidates, evaluate their performance in a single episode, and have human experts review the rendered videos. Specifically, the episode return alone may not reliably indicate the policy's quality.
For instance, in the \texttt{Hopper-v4} environment, achieving a stable standing position yields an episode return of $1000$ but represents a suboptimal strategy as the agent fails to move forward. In contrast, an unstable forward motion might approach the optimal strategy, albeit with a lower episode return.
Thus, human experts evaluate video renderings to assess qualitative aspects of behavior, such as forward progression and stability, to identify the best-performing policy.
Figure~\ref{fig:llm_output} shows an example of a Python policy generated by the LLM, which uses hard-coded proportional-derivative control logic to compute torque actions for basic stability.

\subsection{Masking-Aware TD3}

Previous works, such as \citep{krasowski2023provably, stolz2024excluding}, proposed deterministic action masking in continuous action spaces.
These methods redefine the action space by strictly excluding invalid actions, effectively improving the learning efficiency of RL agents.
However, they rely on the assumption that the mask is fully deterministic, which limits their applicability to scenarios where prior knowledge only suggests that certain actions are suboptimal but not strictly invalid.
This is because such prior information often lacks precise boundaries for defining optimality.

To address this limitation, we propose Masking-Aware TD3, which incorporates dynamic and stochastic masking.
Our approach introduces a probabilistic $\epsilon$-masking mechanism, allowing the RL agent to learn under both masked and unmasked conditions. Specifically, the actor model takes the state $s$ and dynamically computed action bounds $a_{lb} = \pi_{\text{LLM}}(s) - \text{bias}$ and $a_{ub} = \pi_{\text{LLM}}(s) + \text{bias}$.
The actor outputs $x \in [0, 1]$, which is mapped to the constrained action space using $a = \textsc{actionMapping}(x, a_{lb}, a_{ub})$.

By combining guidance from the LLM with stochastic masking, Masking-Aware TD3 enables efficient exploration of the action space while avoiding over-reliance on suboptimal prior guidance.
This approach significantly improves the agent's adaptability and performance across diverse environments.

\subsection{Epsilon Masking}

Epsilon Masking introduces a mechanism to gradually reduce the influence of $\pi_{\text{LLM}}$ over training. Initially, the masking probability $\epsilon_t$ is set to $1$, applying strict constraints based on the LLM’s output.
Over time, $\epsilon_t$ decreases linearly as $\epsilon_t = \max(1 - \frac{t}{f_m \cdot T}, 0.0)$, where $f_m$ is the masking fraction and $T$ is the total duration of training.
This phased reduction enables the RL agent to transition from guided exploration to independent policy learning, optimizing its performance without over-reliance on suboptimal guidance.

\noindent
\begin{minipage}[t]{0.48\textwidth}
\begin{Verbatim}[frame=single, breaklines=true, fontsize=\scriptsize]
Develop a hard-coded policy for the `Hopper-v4` environment using only NumPy within a Python function named `policy(obs)`. The function should map the 11-dimensional observation array `obs` to a 3-dimensional action array `act`.

**Suggested Steps:**

1. **Analyze the Observation Space:** Understand the meaning of each element in the `obs` array.
2. **Define Base Behavior:** Implement logic to encourage forward movement (e.g., applying positive torque to thigh joints).
3. **Implement Balance Control:** Create rules to adjust actions based on the torso angle and other relevant observations to maintain balance.
4. **Refine and Tune:** Experiment with different rules and thresholds to improve the policy's performance.
5. **Ensure Valid Actions:** Use `np.clip` to keep the `act` values within the [-1, 1] range.

<env_infos>{mujoco_xml} {env_infos}</env_infos>

Let's think step by step.
\end{Verbatim}
\captionof{figure}{Prompt for generating a Python policy.}
\label{fig:llm_prompt}
\end{minipage}
\hfill
\begin{minipage}[t]{0.48\textwidth}
\vspace{-0.3em}
\begin{lstlisting}[language=Python]
import numpy as np

def policy(obs):    
    # Extract relevant observations
    z_pos = obs[0]
    ...
    # Define desired state
    desired_z_pos = 1.3
    ...
    # PD gains
    kp_z = 10.0
    ...
    # Calculate control signals (torques)
    torque_thigh = (kp_thigh * (desired_thigh_angle - thigh_angle) - kd_thigh * thigh_angular_vel) + (kp_x_vel * (desired_x_vel - x_vel) - kd_x_vel * torso_angular_vel)
    ...
    # Z-position and torso angle control for basic stability
    torque_foot -= (kp_torso * torso_angle - kd_torso * torso_angular_vel)/4.0
    ...
    return np.clip(np.array([torque_thigh, torque_leg, torque_foot]), -1.0, 1.0)
\end{lstlisting}
\captionof{figure}{Example Python policy generated by LLM.}
\label{fig:llm_output}
\end{minipage}

\section{Experiments}

In this section, we designed three groups of experiments.
First, we evaluated the performance of \textsc{camel}-TD3 under the guidance of expert policies and random policies.
Subsequently, we analyzed the results of experiments with $\pi_{\text{LLM}}$, assessing their potential and characteristics in providing effective guidance.

\subsection{Setup}

Our implementation is based on the PyTorch TD3 from CleanRL~\citep{cleanrl}, with consistent hyperparameters. Key settings include \texttt{Epsilon Masking} ($f_m=0.2$) and \texttt{Action Mapping} ($\text{bias}=0.3$).
For expert masking experiments, we utilize pre-trained models as expert models from CleanRL's HuggingFace repository:
\href{https://huggingface.co/cleanrl/Hopper-v4-td3_continuous_action-seed1}{\footnotesize\texttt{Hopper-v4($3244.59 \pm 8.55$)}}, 
\href{https://huggingface.co/cleanrl/Walker2d-v4-td3_continuous_action-seed1}{\footnotesize\texttt{Walker2d-v4($3964.51 \pm 9.70$)}}, and 
\href{https://huggingface.co/cleanrl/Ant-v4-td3_continuous_action-seed1}{\footnotesize\texttt{Ant-v4($5240.79 \pm 730.24$)}}.

We use the Google Gemini 2.0 (\texttt{gemini-exp-1206}) for generating policies.
All related code, prompts, outputs, and rendered videos are available at \href{https://github.com/sdpkjc/camel-rl}{\texttt{https://github.com/sdpkjc/camel-rl}}.

\begin{figure}[ht]
    \centering
    \includegraphics[width=1.0\linewidth]{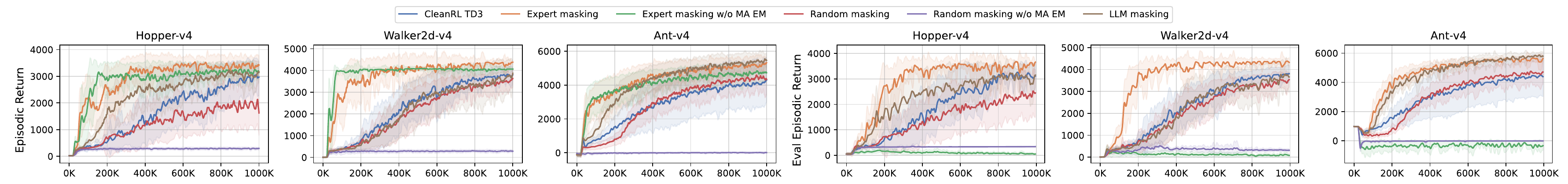}
    \begin{subfigure}[t]{0.48\linewidth}
        \centering
        \vspace{-1em}
        \caption{}
    \end{subfigure}
    \hfill
    \begin{subfigure}[t]{0.48\linewidth}
        \centering
        \vspace{-1em}
        \caption{}
    \end{subfigure}
    \vspace{-0.5em}
    \caption{Episodic return over the time steps for (a) training and (b) evaluation. The shaded area shows one standard deviation over $10$ random seeds. Both curves are smoothed using a rolling average with a window size of $100$. Evaluation curves in (b) are computed every $1000$ timesteps by running three episodes without action masking.}
    \label{fig:camel_curves}
\end{figure}

\subsection{Analysis}


Figure~\ref{fig:camel_curves} shows that applying Masking-Aware (MA) and Epsilon Masking (EM) enables the RL agent to improve returns, even in evaluation environments without expert masking, as epsilon decreases.
In contrast, the control group (Expert masking w/o MA EM) relies entirely on expert masking. In random masking experiments, \textsc{camel} achieves final returns close to the baseline despite random guidance, whereas the control group (Random masking w/o MA EM) fails to learn effective strategies.
These results highlight \textsc{camel}’s ability to utilize expert masking effectively and adapt to random masking conditions.

Rendered videos highlight significant differences in LLM policy performance: continuous hopping in \texttt{Hopper-v4}, standing or tilting in \texttt{Walker2d-v4}, and fast but random walking in \texttt{Ant-v4}. Among five candidate policies generated for each environment, we selected one based on alignment with task objectives. The chosen policies achieved returns of $408.26$ in \texttt{Hopper-v4} (highest return: $1006.95$), $224.70$ in \texttt{Walker2d-v4} (highest return: $1020.58$), and $382.03$ in \texttt{Ant-v4} (highest return: $612.22$).
The varying performance of LLM policies across environments directly impacts \textsc{camel}’s effectiveness. In \texttt{Ant-v4}, \textsc{camel} performs close to expert masking, while in \texttt{Hopper-v4}, it lies between expert masking and baseline. In \texttt{Walker2d-v4}, where the LLM struggles to model bipedal gait, \textsc{camel} performs near baseline, highlighting the importance of initial policy quality for RL training outcomes.

\section{Conclusion, Limitations, and Future Work}

In this work, we introduced the \textsc{camel} framework, which harnesses the capabilities of LLMs to improve RL in continuous action spaces.
By using LLM-generated suboptimal policies as initial guidance and dynamically constraining the action space, \textsc{camel} enhances exploration efficiency and mitigates convergence to suboptimal solutions.
While the approach shows significant promise in environments like \texttt{Hopper-v4} and \texttt{Ant-v4}, its applicability is limited to vectorized observation spaces, and the reliance on expert screening for policy evaluation introduces a manual overhead.
Furthermore, the effectiveness of the framework depends on the underlying LLM’s capability to model complex dynamics, which can be a limiting factor in environments with high dimensionality or intricate task requirements.
Future work could aim to generalize \textsc{camel} to diverse observation and action spaces, automate policy selection to reduce human intervention, and integrate more advanced multimodal LLMs to further enhance adaptability and performance across complex RL scenarios.

\bibliography{rldm}
\bibliographystyle{rlj}

\end{document}